# Decoding Multilingual Topic Dynamics and Trend Identification through ARIMA Time Series Analysis on Social Networks: A Novel Data Translation Framework Enhanced by LDA/HDP Models

*Samawel Jaballi[1,2,3], Azer Mahjoubi[1,2], Manar Joundy Hazar[1,2,4], Salah Zrigui[5], Henri Nicolas[3] and Mounir Zrigui [1,2]*

[1] Faculty of Sciences of Monastir, Univ. of Monastir, Monastir, Tunisia
[2] Research Laboratory in Algebra Numbers theory and Intelligent Systems, Monsastir, Tunisia
[3] Univ. Bordeaux, CNRS, Bordeaux INP, LaBRI, UMR 5800, F-33400 Talence, France
[4] Computer Center, University of Al -Qadisiyah, Qadisiyah, Iraq
[e] Laboratory LIG, CS 40700, 38058 Cedex Grenoble, France

**Abstract.** In this study, the authors present a novel methodology adept at decoding multilingual topic dynamics and identifying communication trends during crises. We focus on dialogues within Tunisian social networks during the Coronavirus Pandemic and other notable themes like sports and politics. We start by aggregating a varied multilingual corpus of comments relevant to these subjects. This dataset undergoes rigorous refinement during data preprocessing. We then introduce our No-English-to-English Machine Translation approach to handle linguistic differences. Empirical tests of this method showed high accuracy and F1 scores, highlighting its suitability for linguistically coherent tasks. Delving deeper, advanced modeling techniques, specifically LDA and HDP models are employed to extract pertinent topics from the translated content. This leads to applying ARIMA time series analysis to decode evolving topic trends. Applying our method to a multilingual Tunisian dataset, we effectively identified key topics mirroring public sentiment. Such insights prove vital for organizations and governments striving to understand public perspectives during crises. Compared to standard approaches, our model outperforms, as confirmed by metrics like Coherence Score, U-mass, and Topic Coherence. Additionally, an in-depth assessment of the identified topics revealed notable thematic shifts in discussions, with our trends identification indicating impressive accuracy, backed by RMSE-based analysis.

*Keywords: Pandemic Governance; Social Media; Multilingual Topic Modeling; Trends Identification; Data Translation; Thematic Patterns; ARIMA; Time Series Analysis*

# 1. Introduction

## 1.1. Background and Motivation

During the Coronavirus pandemic, social media platforms have become the most used channels for instant information exchange and public communication. However, the diverse linguistic environment within social media introduces challenges in crisis communication.
The traditional approaches for topic modeling, which have primarily focused on monolingual data, poses a challenge for a comprehensive analysis of multilingual social media content. This article directly addresses this challenge by proposing a data-driven methodology for multilingual topic modeling. Our proposed approach involves translating multilingual datasets into a single language, upon which topic modeling is applied. Further, we develop a strategy for identifying the trending patterns associated with each topic within the broader context of crisis communication.

## 1.2. Research Objectives

The principal objective of our research is to formulate a proficient data translation framework capable of addressing the intricate multilingual landscape and solving the complexity of multilingual nature prevalent in Tunisian social networks. Additionally, through the integration of Latent Dirichlet Allocation (LDA), Hierarchical Dirichlet Process (HDP) models, and AutoRegressive Integrated Moving Average (ARIMA) time series analysis our goal is to reveal the latent topics and trends within Tunisian textual exchanges throughout the dissemination of the Coronavirus pandemic.

## 1.3. Contribution of the Study

Our research objectives encompass several key facets. Firstly, we survey previous works related to addressing multilingual challenges, setting the foundation for our approach. We endeavor to construct an efficient data translation methodology designed to surmount language barriers, facilitating a more profound analysis of multilingual textual data extracted from Tunisian social networks. An integral aspect involves the assessment and juxtaposition of this methodology against widely adopted techniques, including machine translation techniques, bilingual dictionaries, and crowd-sourced translations. Additionally, we aim to determine the optimal parameters for our LDA and HDP models that effectively handle the nuances of translated dataset, resulting in precise topic modeling. This pursuit is instrumental in attaining an optimal analysis of the presented topics, as well as comprehending and analyzing the intricate dynamics of information flow and public sentiment related specially to the pandemic. The research trajectory also incorporates the application of ARIMA time series analysis, enabling us to decrypt evolving trends linked to the emergent topics. Concluding our endeavor, we undertake an evaluative overview of the results from our translation methodology, the multilingual topic modeling strategy, and our trends identification approach.

## 2. Related Works

### 2.1. Existing Approaches for Multilingual Topic Modeling

Monolingual topic models are trained using a corpus of one single language. Despite their high efficiency and straightforwardness, this methodology proves insufficient in capturing cross-lingual topics. Conversely, multilingual topic models are trained through a corpus comprising text in diverse languages. While this approach accommodates cross-lingual topics, it is more intricate and computationally demanding compared to the simplicity of monolingual topic models [1]. Additionally, in the realm of cross-lingual topic models, an alternative technique involves training on a singular language's text corpus and subsequently extrapolating topics into another language [2]. This method, while exhibiting diminished complexity and computational load in contrast to multilingual topic models, does entail some compromise in terms of precision.

Research by Yang, Boyd-Graber and Resnik in 2019 delves into the effectiveness of various topic modeling methods across different language contexts, offering insights into their strengths and limitations in capturing cross-lingual themes [3]. In their work they have introduced a method for learning weighted topic links across corpora that may have low comparability. Additionally, a comprehensive examination by [4] investigates the intricate balance between computational efficiency and precision within the realm of multilingual topic modeling, providing a comprehensive comparison to monolingual counterparts. They have explored multilingual topic modeling techniques to track COVID-19 trends using Facebook data. Similarly, another study by Xie et al. in 2020 delves into the complex dynamics of cross-lingual topic modeling, they have presented a novel approach to examine topic evolution by leveraging multilingual BERT embeddings and the LDA topic model for both monolingual and multilingual scenarios, leading to a more authentic representation of topics [5].

### 2.2. Topic Modeling in Crisis Communication

In 2020, Boon-Itt and Skunkan harnessed topic modeling methods to delve into tweets revolving around a significant COVID-19 Pandemic event. Their goal was to untangle communication patterns during this crisis [6]. Their analysis identified recurring themes in the tweets, highlighting discussions on the personal effects of the crisis, government response strategies, and media updates. In a separate study published in 2020, Chew and Eysenbach, employed topic modeling techniques to dissect a dataset of tweets pertaining to the H1N1 pandemic. Their analysis brought to light prominent themes in these tweets, encompassing matters such as the dynamics of disease transmission, governmental reactions to the crisis, and the socio-economic disruptions the pandemic triggered [7]. Additionally, Wang et al. introduced an innovative approach to topic modeling aimed at extracting essential information from social media during a crisis. Researchers who adopted this approach examined a dataset of crisis-related tweets, successfully identifying vital information that the public sought amidst the ongoing crisis [8].

## 2.3. Machine Translation Techniques

This subsection provides an in-depth exploration of pertinent research centered on prominent machine translation systems. The discussion concludes with a comprehensive overview of the broader implications and insights derived from the various machine translation techniques.

### 2.3.1. Rule-Based Machine Translation

An early approach to language translation is Rule-Based Machine Translation (RBMT), which is based on linguistic rules and dictionaries. It entails the manual creation of linguistic rules that deal with word translations. According to Khanna et al. [9], these rules cover grammar, semantics, and syntax. Complex linguistic nuances and idiomatic expressions present challenges due to variability, context, and rule limitations. They are generally labor-intensive to develop and maintain, requiring experts for ongoing rule set maintenance as languages evolve. Difficulties arise for RBMT when facing intricate linguistic structures and idiomatic sentences that diverge from grammatical norms. Moreover, language evolution necessitates frequent maintenance to uphold translation accuracy. However, this procedure proves demanding in terms of resources and slow in terms of adaptation due to the need for linguistic and domain experts for rule set maintenance.

### 2.3.2. Statistical Machine Translation

Statistical machine translation (SMT) operates as a data-driven method, relying on extensive bilingual corpora to identify translation patterns and probabilities. According to Koehn et al. [10], these systems identify accurate translations based on observed patterns in training data. A popular method within SMT is phrase-based translation, where translations are learned for chunks of words rather than individual words. SMT systems can produce accurate translations, but they encounter difficulties in capturing long-range dependencies and handling rare or unseen phrases. The method's limitations include capturing extensive sentence dependencies that stretch over a significant distance, and the inability to understand idiomatic phrases and linguistic nuances, potentially leading to less accurate translations.

### 2.3.3. Neural Machine Translation

Neural machine translation is a recent paradigm garnering significant attention and producing accurate results. Vaswani et al. [11] have reviewed methods, resources, and tools of neural machine translation. NMT utilizes artificial neural networks, particularly recurrent or transformer models, to translate texts between languages. NMT models excel in capturing complex linguistic structures and nuances due to extensive training on large parallel corpora. They adeptly handle long-range dependencies, and the attention mechanism enhances the alignment of source and target language words. NMTs offer customization options, including domain-specific terminology specification. Despite improved translation quality, NMT systems demand substantial computational resources, especially for real-time translations.

### 2.3.4. Hybrid Machine Translation

Hybrid machine translation (HMT) combines various approaches to overcome limitations of individual methods. It generally integrates rule-based, statistical, and/or neural machine translation techniques. For instance, a hybrid system might use a statistical model for initial translation and then apply rule-based post-editing for refinement. The goal is to combine different strengths for more accurate translations. In [12], the combination of different methods poses challenges in terms of synchronization and harmonious blending, which can lead to complexities within hybrid systems. Additionally, due to the combination of different modules, the maintenance of the parts can be an overhead burden.

2.3.5. Discussion

In summary, each machine translation approach presents distinct strengths and challenges. RBMT establishes a robust foundation but grapples with adaptability, while hybrid methods leverage strengths amidst navigational complexities. On the other hand, SMT boasts commendable accuracy, yet nuanced linguistic elements pose hurdles. Meanwhile, NMT emerges as a cutting-edge solution, although its resource-intensive nature warrants consideration. The approach chosen hinges on specific translation task requirements and constraints. In this context, the NMT emerges as a superior and high-performing approach, demonstrating its proficiency in effectively handling intricate nuances, linguistic structures, and even long-range dependencies. Nonetheless, it's important to carefully consider its resource-intensive nature in relation to the translation task's demands and the available computational capacity. Furthermore, the scalability of NMT systems offers a distinct advantage, enabling efficient management of substantial translation tasks, without the necessity for direct hardware resource management [13]. While they may not provide an equivalent level of model customization compared to certain open-source alternatives, these systems do offer customization options. This empowers us to define domain-specific terminology, thereby meeting specific requirements with precision.

## 2.4. Multilingual Text Processing in Social Media Analysis

A survey conducted by [14] offers an extensive exploration of the intricacies and obstacles associated with processing multilingual text on social media platforms. This study delves into the use of advanced deep learning techniques for analysing sentiments across different languages. It explores the challenges and opportunities in this domain, shedding light on the application of deep learning in multilingual sentiment analysis and language identification. Similarly, another survey authored by Hazar et al. [15], provides a comprehensive overview of different methodologies within the realm of multilingual topic modeling and. The authors extensively outline the primary challenges associated with multilingual open information extraction, emphasizing the linguistic diversity, structural variations, the lack of comparable corpora and the scarcity of resources in certain languages. Additionally, Jain et al. [16] delved deep into the realm of extracting emotions from multilingual text, particularly from the dynamic landscape of social media. Examining user reactions across various domains such as political elections, sports and healthcare, their proposed framework incorporated linguistic, psychological, and emotion theories. This intricate design ensured a comprehensive understanding of emotion lexicons, setting it apart from conventional sentiment analysis techniques. Their findings highlighted the model's effectiveness in accurately identifying cross-lingual events, a task that monolingual topic models had not achieved.

## 3. Proposed Methodology

Our approach is structured into five primary phases, as depicted in Figure 1. Firstly, the data collection phase involves gathering and constructing a corpus of multilingual pandemic-related comments. Following this, the data preprocessing phase focuses on preparing the collected data. Thirdly, we introduce our No English-English Machine translation approach. The fourth phase involves topic modeling, where we leverage LDA and HDP models to extract meaningful topics from the translated data. Lastly, the final phase revolves around trends identification, where we employ time series analysis based on ARIMA model to uncover trends within the extracted topics.

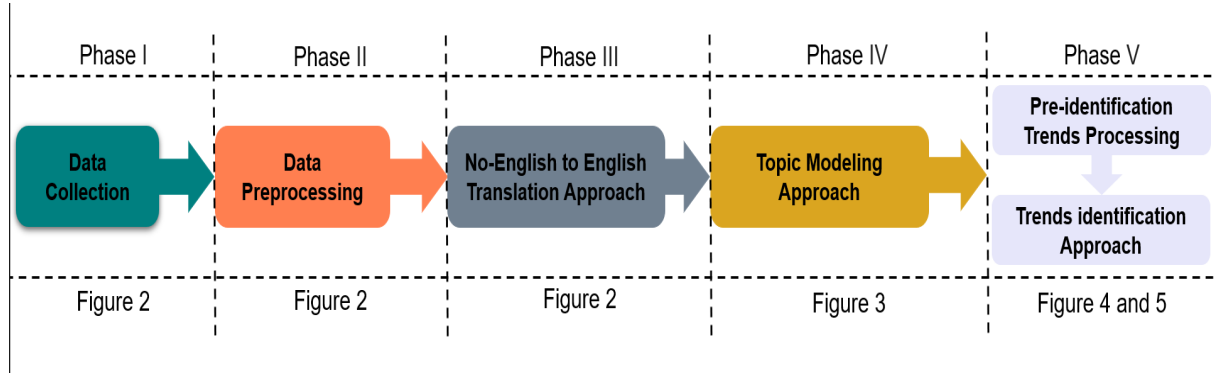

*Figure 1. The general architecture of our methodology*

### 3.1. Data Collection Phase

The data collection strategy for this study involved incorporating multiple corpora, comprising a combination of Arabic and Latin texts, as elaborated below: The ASTD corpus [17], which aggregates 30K pandemic-related texts sourced from official channels. Notably, the texts are exclusively in Modern Standard Arabic (MSA). Complementing this, the CSTA corpus [18], encompassing 25K texts in Arabizi and 15K in Dialectical Arabic, was assembled during the pandemic from discussions occurring on platforms such as the official Facebook page of the Tunisian Minister of Health. Lastly, the TUNIZI corpus [19] comprises a compilation of 60K Arabizi texts and 30K Latin texts including French and English extracted from YouTube channels and Twitter pages of Tunisian Parliament records, as well as pandemic news, facilitating an in-depth exploration of linguistic variations. Table 1 provides insights into the collected corpus, and table 2 highlights the distribution of languages within the corpus, providing a focused perspective on the linguistic composition of the dataset.

*Table 1. Corpus Statistics*

| #Text rows | #Sentences | #words | #unique words |
|---|---|---|---|
| 160,000 | 238,567 | 4,567,890 | 89,234 |

*Table 2. Language Distribution Statistics*

| % of Arabic | % of French | % of Arabizi | % of English |
|---|---|---|---|
| 36 | 23 | 19 | 22 |

## 3.2. Data Preprocessing Phase

To preserve the semantic meaning of the text, we applied the pre-processing steps inspired by Jaballi et al. [20] to the collected dataset. These steps include removing punctuation marks, such as periods, commas, question marks, etc., as they often do not carry significant meaning for translation and topic modeling tasks. Additionally, special characters, symbols, and numeric digits that do not hold significant meaning for translation and topic modeling tasks were removed. Furthermore, we replaced numbers existing in letters with their corresponding substitute letters, based on their morphological resemblances to Arabic letters such as the case of the Arabizi words (e.g., 7 => h; n7ebek => nhebek) Furthermore, all text formats were converted to lowercase to ensure consistency and to prevent the treatment of words with different cases as separate entities. Tokenization was performed as well, dividing the text into individual tokens. This process helps break down the multilingual text into smaller units. Concurrently, we conducted the removal of stopwords to eliminate noise and to further enhance the text for analysis. In the translated experiment, we excluded common low-information words that do not contribute significantly to the meaning of the text, such as (e.g., "par", "sur" "is", "على", "often", "إلى", etc.) Additionally, we employed lemmatization techniques to transform words into their base form such as converting ("better" to "good") This process contributes to text normalization and diminishes word variations. As well, applying stemming involves truncating words to their root, for instance, changing ("running" to "run") and ("إنتشار" to "نشر") Thus, table 3 showcases the statistics of the text data before and after each preprocessing step.

Table 3. Comparison Of Text Characteristics Pre- and Post-Preprocessing

| Preprocessing Step | Before Preprocessing | After Preprocessing |
|---|---|---|
| Punctuation and Symbols Removal, | 4,567,890 words | 4,354,209 words |
| Numbers and Digits Removal | 4,354,209 words | 4,107,216 words |
| Stopwords Removal | 4,107,216 words | 4,000,831 words |
| Lemmatization | 89,234 unique words | 87,739 unique words |

## 3.3. No English-English Machine Translation approach

Our machine translation approach, depicted in Figure 2, capitalizes on the integration of two Bilingual dictionaries and an Arabic lexicon, facilitating connection and correspondence through a parallel dictionary. Subsequently, a validation phase is employed using crowd-sourced translation via the PROZ platform. This is followed by the evaluation of the translation set, employing two key metrics: Accuracy and F1-Score. Furthermore, the final step involves translation through the Open AI API, with evaluation conducted after each iteration, subsequently leading to incremental improvements in accuracy. In the upcoming subsections, we will provide in-depth explanations of each step, offering comprehensive insights into the intricacies of our approach.

### 3.3.1. Bilingual Dictionaries and Arabic Lexicon

Bilingual dictionaries and lexicons play a pivotal role in machine translation systems, providing words and phrases with corresponding English translations, enhancing translation accuracy and precision. These resources fall into two main categories: parallel dictionaries, facilitating accurate translation alignment between source and target languages, and the second category is comparable dictionaries, which capture similar meanings in French, Arabic, and English [21]. As well, Arabic lexicons, exclusively containing words, aid in understanding Arabic context and refining translation accuracy.

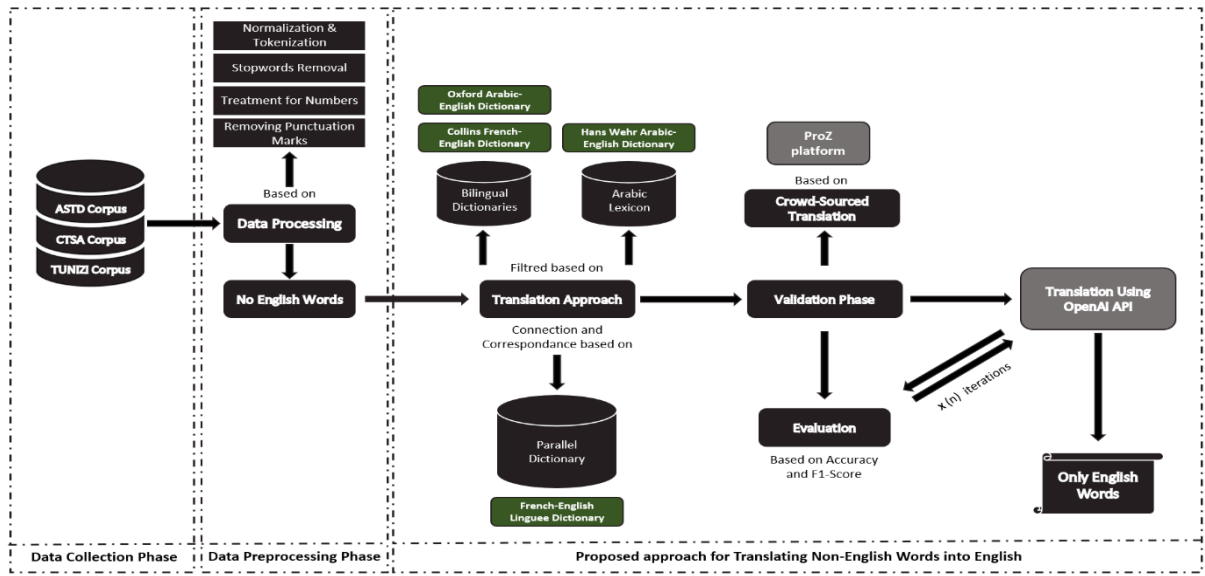

*Figure 2. The general architecture of the Machine Translation system*

In the process of integrating linguistic resources for our translation approach, we have leveraged two essential bilingual references, namely the "Collins French-English Dictionary" was conducted by [22] and the third volume of "Oxford Arabic-English Dictionary" edited by [23], in conjunction with the comprehensive "Hans Wehr Arabic-English Dictionary" by [24] as an Arabic lexicon. These resources have collectively served as the backbone of our translation process, enabling us to convert non-English content into English by effectively mapping words and phrases from French and Arabic to their English equivalents. Through these references, we have methodically mapped words and expressions from both French and Arabic, effectively capturing their respective meanings and contexts.

Central to our approach is the strategic establishment of connections and correspondence among linguistic elements. This pivotal task is facilitated by a "French-English Linguee Parallel Dictionary", a dynamic online resource that serves as a direct correspondence between French terms and their English equivalents. By facilitating precise translation and alignment of meaning between the two languages, this parallel dictionary bolsters the accuracy and contextual fidelity of our translations, ensuring precise alignment of linguistic elements and enhancing translation accuracy for nuanced meanings and effective cross-language communication.

### 3.3.2. 1st validation stage and Crowd-Sourced Translation

The initial validation phase of our machine translation approach involves a pivotal step known as "Crowd-Sourced Translation", facilitated through the PROZ platform. During this stage, the collaboration of skilled human translators becomes instrumental as they meticulously review, enhance, and refine the machine-generated translations. This collaborative effort aims to address linguistic nuances, cultural context, and domain-specific intricacies that might be overlooked by automated systems [25]. Upon completing the initial machine-generated translations based on Bilingual dictionaries and Arabic lexicon, they undergo a thorough review by human translators on the PROZ platform. Drawing upon their profound grasp of both the source and target languages, these translators are adept at identifying any discrepancies, ambiguities, or potential inaccuracies introduced during the translation process. Through iterative cycles of feedback and revision, the translations undergo refinement, with a focus on enhancing their contextual accuracy and linguistic precision. Nevertheless, this phase is particularly sensitive to the nuances and potential errors inherent in translating Arabizi words in English. These forms of expression carry specific meanings tailored to the Tunisian population, which may pose unique challenges.

### 3.3.3. Enhancing Translation Performance with OpenAI API Integration and In-Depth Evaluation

In this phase of our methodology, we delve into the intricacies of harnessing the OpenAI API to enhance the translation, representing a significant advancement in our approach. The strategic integration of the OpenAI API complements our earlier stages, notably the incorporation of Bilingual dictionaries and the Arabic lexicon. Operating as a powerful tool, the OpenAI API augments our translation capabilities through its robust neural network architecture. When the API processes source language input, it generates translated text in the target language, leveraging its extensive linguistic comprehension and adept translation proficiency. However, translating non-English words, especially from the Arabizi languages, presents a substantial challenge. To address this, the OpenAI API leverages its advanced neural network model, meticulously trained on a diverse multilingual dataset. This training equips the model to capture intricate linguistic relationships and nuances across languages, enabling it to adeptly map non-English words to their English equivalents. The contextual understanding embedded within the neural network assists in determining the most suitable English translation, accounting for the contextual nuances of the sentence. This meticulous consideration ensures translations that are not only accurate but also contextually meaningful.

The philosophy underpinning the OpenAI API's translation model is rooted in GPT-3, utilizing a decoder-only transformer architecture. This architectural choice draws upon neural network design principles and plays a pivotal role in enabling advanced translation capabilities. The decoder-only transformer architecture, tailored for sequence-to-sequence tasks like translation, proves to be a potent neural network variant. At its core, the decoder-only transformer architecture consists of multiple layers incorporating self-attention mechanisms and feedforward neural networks [26]. These layers collaborate to process input sequences, capturing intricate patterns and interdependencies within the data. The self-attention mechanisms empower the model to discern the significance of individual words within a sentence, considering contextual cues and semantic nuances. This facet holds particular significance in translation tasks, where a profound understanding of the source language's context and intricacies is essential for generating precise and meaningful translations. Mathematically, we can represent the decoder-only transformer architecture as follows:

$$X = [x_1, x_2, \ldots, x_n] \wedge H_i = Layer(H_{i-1}) \; \forall \; 1 \leq i \leq L \tag{1}$$

$$E = Embeddings(X) + PositionalEncoding(X) \tag{2}$$

$$H_i = SelfAttention(H_{i-1}) + FeedForward(H_{i-1}) \tag{3}$$

$$SelfAttention(H_{i-1}) = Softmax\left(\frac{H_{i-1}W_Q(H_{i-1}W_K)^T}{\sqrt{d_k}}\right) H_{i-1} W_V \tag{4}$$

$$FeedForward(H_{i-1}) = ReLU(H_{i-1}W_1 + b_1)W_2 + b_2 \tag{5}$$

$$Y = Softmax(W_o H_L + b_o) \tag{6}$$

In the context of sequence-to-sequence translation, an input sequence $X$ composed of words in the source language is processed through an architecture consisting of $L$ layers denoted as $H_i$ for $1 \leq i \leq L$. Initially, the input sequence $X$ is transformed into an embedded form, enhanced with positional encodings to preserve the sequence order. Each layer's processing is defined by the equation (3), where $H_i$ represents the current layer's hidden state and $H_{i-1}$ is the input from the preceding layer. The self-attention mechanism computes attention scores among words in the sequence, enabling the model to emphasize relevant words contextually. Mathematically, this mechanism can be expressed as (4), incorporating learnable weight matrices $W_Q$, $W_K$, and $W_V$ and a dimension parameter $d_k$. The feedforward neural network, defined as (5), transforms the input hidden state through linear transformations with weights $W_1$ and $W_2$ and biases $b_1$ and $b_2$ followed by $ReLU$ activation. The culmination of this process is the generation of the final output sequence $Y$, as represented in equation (6). This output is derived from the last layer's output $H_L$ and involves applying weights $W_o$ and biases $b_o$ before a softmax function to produce the final translation. This architecture captures intricate relationships, making it ideal for tasks like sequence-to-sequence translation, where contextual nuances are vital for accurate output generation.

Subsequently, our evaluation phase involved partitioning the collected corpus into training and testing subsets, employing 70% for training and 30% for testing. The evaluation process itself employs the utilization of the Accuracy and F1-Score metrics. Furthermore, our methodology integrates an iterative translation cycle, meticulously intertwined with simultaneous evaluations. This dynamic interplay leads to discernible enhancements in the performance of the translation strategy and showcases our commitment to achieving optimal results. This iterative process allows us to fine-tune the translation model based on the insights gained from previous translation training and evaluations. With each iteration, the model learns from its past errors and refines its understanding of the linguistic nuances, contextual intricacies, and idiomatic expressions that are characteristic of many expressions in our dataset, enhancing in best accuracy of the translation model. Table 4 highlights the resources and techniques employed in our translation process.

*Table 4. Overview of Employed Resources and Techniques in Translation Approach*

| Resource | Type | Purpose |
| --- | --- | --- |
| Collins French-English Dictionary | Bilingual Dictionary | Mapping French to English |
| Oxford Arabic-English Dictionary | Bilingual Dictionary | Mapping Arabic to English |
| Hans Wehr Arabic-English Dictionary | Arabic Lexicon | Understanding Arabic context |
| French-English Linguee Parallel Dictionary | Parallel Dictionary | Facilitating precise translation and alignment of meaning between French and English |
| PROZ platform | Crowd-Sourced Translation Platform | Review and refine machine-generated translations |
| OpenAI API | Neural Network Model | Enhancing translation of non-English words into English |

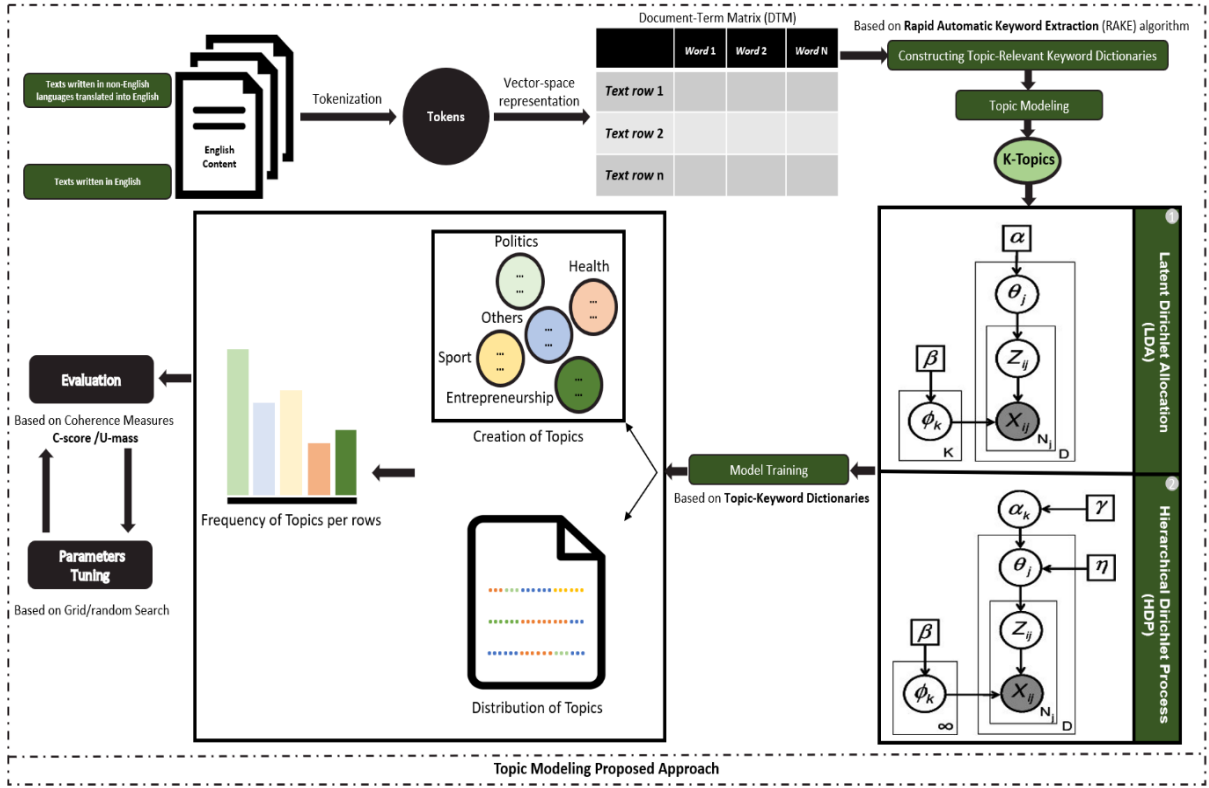

*Figure 3. The general architecture of the Topic Modeling system*

## 3.4. Topic Modeling Proposed Approach

In this section, we outline our proposed methodology for topic modeling. Our approach, as illustrated in Figure 3, involves leveraging LDA and HDP algorithms to extract latent topics from the English translated textual input.

### 3.4.1. Overview and Thematic Insights using LDA & HDP

In the realm of topic modeling, LDA and HDP emerge as cornerstone models, each with its distinct prowess in illuminating the underlying thematic intricacies embedded within extensive collections of documents. LDA, a venerable statistical model, operates on the fundamental assumption that documents can be seen as compositions of a pre-defined number of topics. Words are intricately associated with specific thematic categories [27]. The core concept revolves around representing each topic as a distribution of words, capturing the recurrent co-occurrence of certain terms. This architectural blueprint inherently encapsulates its objective, as visually demonstrated in block number 1 in Figure 3. Furthermore, HDP, an evolutionary extension of LDA, shares a comparable structural foundation. Illustrated in block number 2 in Figure 3, this model exhibits a nested hierarchy of topics. Notably, it adds a dynamic dimension by automatically determining the optimal number of topics ($K^*$) directly from the dataset itself. This feature empowers HDP with a hierarchical architecture that transcends conventional models in terms of flexibility and adaptability. Research at the confluence of LDA and HDP models, along with the exploration of topic dynamics, has seen significant advancement, as evidenced by recent scholarly endeavors. Jelodar et al. [28] made noteworthy contributions by extending the LDA model to encompass dynamic topic modeling, allowing for the temporal evolution of topics within social media data. Another prominent illustration of contemporary research involving HDP models is the work conducted by Lubis et al. [29]. In their study, they delve into the dynamic characteristics of topics present in social media content, introducing a pioneering approach grounded in the application of supervised HDP models.

3.4.2. Mathematical Foundations of LDA and HDP Process Models

Turning our attention to the mathematical foundation, both LDA and HDP are generative probabilistic models tailored to handle discrete data, including bags of words extracted from text documents, commonly referred to as topic models. To illustrate the notation, we refer the reader to the graphical models for LDA and HDP shown in blocks 1 and 2 in Figure 3. Additionally, to enhance readability, we have listed commonly used variables in Table 5.

Table 5. Summary of Used Variables

| variables | Description |
|---|---|
| $D$ | Number of documents in collection |
| $W$ | Number of distinct words in vocabulary |
| $N$ | Total number of words in collection |
| $K$ | Number of topics |
| $x_{ij}$ | $i^{th}$ observed word in document $j$ |
| $z_{ij}$ | Topic assigned to $x_{ij}$ |
| $N_{wk}$ | Count of word assigned to topic |
| $N_{kj}$ | Count of topic assigned in document |
| $\phi_k$ | Probability of word given topic $k$ |
| $\theta_j$ | Probability of topic given document $j$ |

LDA conceptualizes each of the $D$ documents in a collection as a blend across $K$ latent topics. Each topic is represented as a multinomial distribution encompassing a vocabulary of $W$ words. When considering a document indexed as $j$, the process commences by extracting a mixing proportion denoted as $\theta_j$ derived from a Dirichlet distribution with parameter $\alpha$. For the $i^{th}$ word within the same document, a topic assignment, $z_{ij} = k$, is chosen with a probability defined by $\theta_{k|j}$. Consequently, the word $x_{ij}$ is drawn from the corresponding topic $z_{ij}$, with the probability of selecting word $w$ being governed by $\phi_{w|z_{ij}}$. This process is guided by a Dirichlet prior featuring the parameter $\beta$, which influences the characteristics of the word-topic distributions $\phi_k$. Therefore, the generative process for LDA is established as:

$$\theta_j \sim \mathcal{D}[\alpha], \; \phi_k \sim \mathcal{D}[\beta], \; z_{ij} \sim \theta_j, \; x_{ij} \sim \phi_{z_{ij}} \tag{7}$$

To simplify notation while preventing unnecessary complexity, we adopt the abbreviated form $\theta_j \sim \mathcal{D}[\alpha]$ to represent $[\theta_{1|j}, \ldots, \theta_{k|j}] \sim \mathcal{D}[\alpha, \ldots, \alpha]$ for the purpose of Dirichlet sampling, and similarly for $\phi$. The comprehensive joint distribution spanning all parameters and variables is articulated as:

$$p(x, z, \theta, \alpha, \beta) = \prod_j \frac{\Gamma(K\alpha)}{\Gamma(\alpha^K)} \prod_k \theta_{k|j}^{N_{kj}+\alpha-1} \prod_k \frac{\Gamma(W\beta\alpha)}{\Gamma(\beta^W)} \prod_w \phi_{w|k}^{N_{wk}+\beta-1} \tag{8}$$

Here, $N_{wkj}$ signifies the count of occurrences where $x_{ij} = w$ and $z_{ij} = k$, employing the convention of summation over missing indices. The entities $N_{kj} = \sum_w N_{wkj}$ and $N_{wk} = \sum_j N_{wkj}$ are the central count arrays utilized in computations, denoting the number of words allocated to topic $k$ in document $j$ and the occurrences of word $w$ being associated with topic $k$ in the corpus, respectively. Given the observed words $x = \{x_{ij}\}$, the objective in Bayesian inference for LDA involves the calculation of the posterior distribution concerning latent topic assignments $z = \{z_{ij}\}$, the mixing proportions $\theta_j$, and the topics $\phi_k$. Approximate inference for LDA can be undertaken using contemporary variational techniques (e.g., Chemudugunta et al. [30]) or advanced probabilistic inference strategies (e.g., Dubray et al. [31]).

As for the HDP, it constitutes a collection of Dirichlet Processes, sharing identical topic distributions, and can be perceived as the non-parametric extension of LDA. The distinct advantage of HDP is its capacity to derive the number of topics from the available data. The formulation of the HDP model arises when considering the limit as $K$ tends towards infinity [32]. To elaborate, let us introduce $\alpha_k$ as top-level Dirichlet variables, drawn from a Dirichlet distribution with parameter $\gamma/K$. The topic mixture associated with each document, denoted as $\alpha_j$, is extracted from a Dirichlet distribution characterized by parameters $\eta\alpha_k$. The word-topic distributions $\phi_k$ are drawn from a fundamental Dirichlet distribution governed by parameter $\beta$. Similar to LDA, the assignment of topic $z_{ij}$ is drawn from $\theta_j$, and the selection of word $x_{ij}$ stems from the corresponding topic $\phi_{z_{ij}}$. Summarily, the generative process unfolds as follows:

$$\alpha_k \sim \mathcal{D}[\gamma/K], \phi_j \sim \mathcal{D}[\eta\alpha_k], \phi_k \sim \mathcal{D}[\beta], z_{ij} \sim \theta_j, x_{ij} \sim \phi_{z_{ij}} \quad (9)$$

### 3.4.3. Topic Modeling Proposed Framework: From Keyword Extraction to Probabilistic Modeling and Evaluation

In this section, we introduce our designed framework that targets the extraction of latent topics from the English translated data. This endeavor commences with a fundamental step: the tokenization of the text, wherein each document is methodically broken down into individual tokens. This initial tokenization sets the stage for the creation of a Document-Term Matrix, a cornerstone in the process of generating a vectorized representation of documents. Within this matrix, documents are assigned rows, while columns accommodate the unique words present across the entire corpus. This matrix inherently encodes the distribution of term frequencies within documents, laying a robust foundation for the subsequent extraction of latent topics.

To enhance the precision of topic extraction, we employ pertinent keyword dictionaries crafted using the Rapid Automatic Keyword Extraction (RAKE) algorithm. This approach starts by segmenting sentences into individual words, forming a set of potential keyword candidates. Next, the algorithm calculates how often each word appears in the text. It then thoroughly examines how these words co-occur within specific contexts. Based on the insights from these frequencies and co-occurrences, each word is given a score. After ranking the words by their scores, the top-performing keywords are extracted. These selected keywords are then grouped by their meaning and themes, resulting in the creation of comprehensive keyword dictionaries across various domains. These extracted keywords are then organized based on both semantic and thematic similarity, leading to the creation of keyword dictionaries that span diverse domains. Upon amalgamating these keyword dictionaries, the emergence of at least five distinct categories becomes evident. These categories are discerned through meticulous manual assessment based on the extracted keywords. The domains of sports, health, politics, entrepreneurship, and others are comprehensively covered within these categories. It's worth highlighting that this phase can be seen as a "Pre-annotation" step, offering an initial glimpse into the semantic distribution of our data before applying topic modeling algorithms. Subsequently, we initialize the number of desired topics, referred to as "k-topics," to five, aligning with the identified categories.

As we progress, our pipeline transitions into the realm of probabilistic modeling. First, the LDA operates as a cornerstone model within the domain of topic modeling, adeptly capturing intricate relationships between words and documents within a corpus. This method views documents as compositions of underlying topics, each characterized by a distribution of words. The chief strength of LDA lies in its ability to uncover these latent topics and their associated word distributions, thereby unveiling the underlying themes within the dataset. Simultaneously, the LDA algorithm is parameterized by two crucial hyperparameters: $\alpha$ (alpha) and $\beta$ (beta). The $\alpha$ parameter shapes the document-to-topic distribution, determining how documents consist of various topics. In contrast, the $\beta$ parameter influences the topic-to-word distribution, impacting how topics are defined by sets of words. The choice of these hyperparameters significantly influences the outcomes of the LDA model. Effective strategies for optimizing these parameters encompass techniques such as cross-validation or probabilistic graphical methods. The process of LDA entails iteratively assigning topics to words and documents while updating the topic-word and document-topic distributions. Moreover, Gibbs Sampling is employed for this iterative assignment, facilitating convergence to meaningful topic distributions. In comparison, we also employed HDP which extends LDA's capabilities by offering a more adaptable approach to topic modeling. Unlike LDA, which requires a predetermined number of topics, HDP can autonomously determine the number of topics from the data, adapting to the inherent complexity of our dataset. As well, HDP employs a nested hierarchy of topics, empowering the model to encompass a broader spectrum of topic relationships. This proves particularly advantageous when dealing with datasets characterized by overlapping themes or subtopics [33].

The final outcome encompasses a coherent assembly of distinct topics, each meticulously characterized by its unique distribution of words. Furthermore, this strategic allocation of topics significantly enriches the overall comprehension of the distribution of content within the documents. To enhance the depth and pertinence of the derived topics, an in-depth statistical analysis was performed on the five identified categories. Table 6 encapsulates the average word frequency, the prevalence of each topic, and the distribution of documents across these five pivotal topics.

*Table 6.      Statistical Summary of Extracted Topics*

| Category | Average Word Frequency | Document Allocation | Top Keywords |
|---|---|---|---|
| Health | 0.0228 | 31% | COVID-19, Patient, Prevention |
| Sports | 0.0176 | 24% | Game, Team, Competition |
| Politics | 0.0153 | 21% | Government, Election, Vote |
| Entrepreneurship | 0.0119 | 16% | Business, Strategy, Startup |
| Others | 0.0060 | 08% | Fashion, Travel, Trend |

During our evaluation process, we have employed the U-Mass and Coherence Score, a pivotal aspect of our assessment framework, to gain insights into the semantic coherence of topics. The U-Mass score, represented by the equation:

$$\text{UMass} = \left(\frac{1}{N(N-1)}\right) * \sum \left[log\left(\frac{(D(w_i,w_j)+1)}{D(w_i)}\right)\right] \quad (10)$$

where $i \neq j$ and $N$ is the number of unique words in the topic and $D(w_i, w_j)$ is the number of documents containing both words $w_i$ and $w_j$, offers a quantitative perspective on how word interrelationships can be interpreted. This is achieved by calculating pairwise similarity using Pointwise Mutual Information (PMI), given by the equation:

$$PMI(w_i, w_j) = log^2\left(\frac{P(w_i,w_j)}{(P(w_i)*P(w_j))}\right) \quad (11)$$

where $P(w_i, w_j)$ is the joint probability of words $w_i$ and $w_j$ occurring together, and $P(w_i)$ and $P(w_j)$ are the individual probabilities of words $w_i$ and $w_j$ occurring. Notably, an increase in $UMass$ values signify topics with improved coherence and semantic structure. To enhance our evaluation, we introduce the $C-score$, a refinement over $UMass$. The Coherence Score represented by the equation:

$$C - score = \left(\frac{1}{N(N-1)}\right) * \sum (PMI\ (w_i, w_j)) \quad (12)$$

where $PMI(w_i, w_j)$ is the Pointwise Mutual Information between words $w_i$ and $w_j$, advances coherence measurement, making it more intuitive and interpretable. To effectively utilize these evaluation mechanisms, it is essential to align the topic-word distribution and a meticulously constructed document-term matrix capturing word co-occurrences. However, we emphasize that the effectiveness of our approach relies not only on evaluation but also on parameter adjustments, which form the foundation of our models. As a result, we have commenced with parameter tuning, aiming to identify optimal settings, including Number of Topics ($k$), Alpha ($\alpha$), Beta ($\beta$), and Concentration parameters. To achieve this, we employ optimal parameter tuning techniques including Grid Search and Random Search, where Grid Search explores all potential configurations, and Random Search selects values randomly from predetermined ranges, proving effective in high-dimensional parameter spaces.

### 3.5. Pre-Identification Trends Processing: Decoding Fixed Trends Through Semantic Analysis Supported by Supervised Learning Strategy

Proceeding to the penultimate step, we focus on the pre-identification trends processing, as depicted in Figure 4. Since we commence without any pre-established trends, the key aim of this method is to ultimately define five constant trends. This method functions by converting the trend identification issue into a supervised learning task. To start, we extract the most recurrent words from data related to trends. Preprocessing techniques employing tokenization allow us to effectively manage syntactic variations. Subsequently, lemmatization is applied, utilizing the WordNet Lemmatizer, a linguistic tool that transforms words into their lemmas forms. The WordNet Lemmatizer leverages the extensive WordNet lexical database, a comprehensive resource that links words to their semantic relationships, synonyms, and definitions [34]. As well, WordNet is a large lexical database of English, in which nouns, verbs, adjectives, and adverbs are organized into interlinked synsets. Each synset consists of a set of synonyms representing a single underlying lexical concept. By referencing this rich repository, the WordNet Lemmatizer not only accurately converts words to their root forms but also retains the semantic meanings and relationships between words. This is crucial because many words have different meanings and consequently, different lemmas in different contexts. Thus, the WordNet Lemmatizer takes into account both the word's part of speech and its context to accurately identify the best-suited lemma, guaranteeing not just precision, but also a lemmatization that's rich in meaning. Additionally, we applied the SymSpell tool, which is instrumental in spelling correction, thereby enhancing the overall integrity of the extracted words. SymSpell employs the Symmetric Delete spelling correction algorithm, specifically designed for its speed and efficiency, thereby making it well-suited for real-time applications. This algorithm operates in three interconnected phases. In the preprocessing phase, a dictionary is generated containing all conceivable "deletes", which are words created by removing a single character from every word in the vocabulary. Each of these "deletes" is then stored in the dictionary, along with a pointer linking back to the original word or words from which it could have originated. Moving on to the Lookup phase, the algorithm delves into the dictionary, generating all the possible "deletes" of the misspelled word to be corrected. This results in a compiled list of candidate words that are within an acceptable edit distance from the original, incorrect word. Lastly, in the Ranking phase, these candidate words are meticulously ranked based on set criteria including frequency and edit distance, culminating in the selection of the top-ranked candidate as the corrected word [35]. In the subsequent step, words are converted into numerical vectors using the pre-trained FastText model, a state-of-the-art word embedding methodology developed by Facebook AI Research (FAIR). Fast Text is an open-source library designed to generate word embeddings for text data which are dense vectors representing words in a consistent vector space, essential for feeding text data into machine learning models, as most models require numerical input. Thus, FastText, transcending conventional models like Word2Vec, introduces a novel methodology by integrating subword units, character n-grams, and entire words into its computational framework. This innovative approach allows FastText to capture subtle syntactic and semantic variations in words, which is crucial for processing languages with complex morphologies or those generating new, previously unregistered words, as in our translated dataset. The model's unique word representation method, akin to the CBOW model of Word2Vec but with added sophistication through subword units, enables it to generate word embeddings, understand morphemes, and handle out-of-vocabulary words. During training, the model learns the vector representations of n-grams, aiming to predict words from their context and minimize prediction loss. Subsequently, the trained model generates word embeddings by summing the embeddings of a word's n-grams, proving invaluable for constructing vectors of words absent in the vocabulary. Table 7 encapsulates the Word Embedding Statistics generated by FastText model.

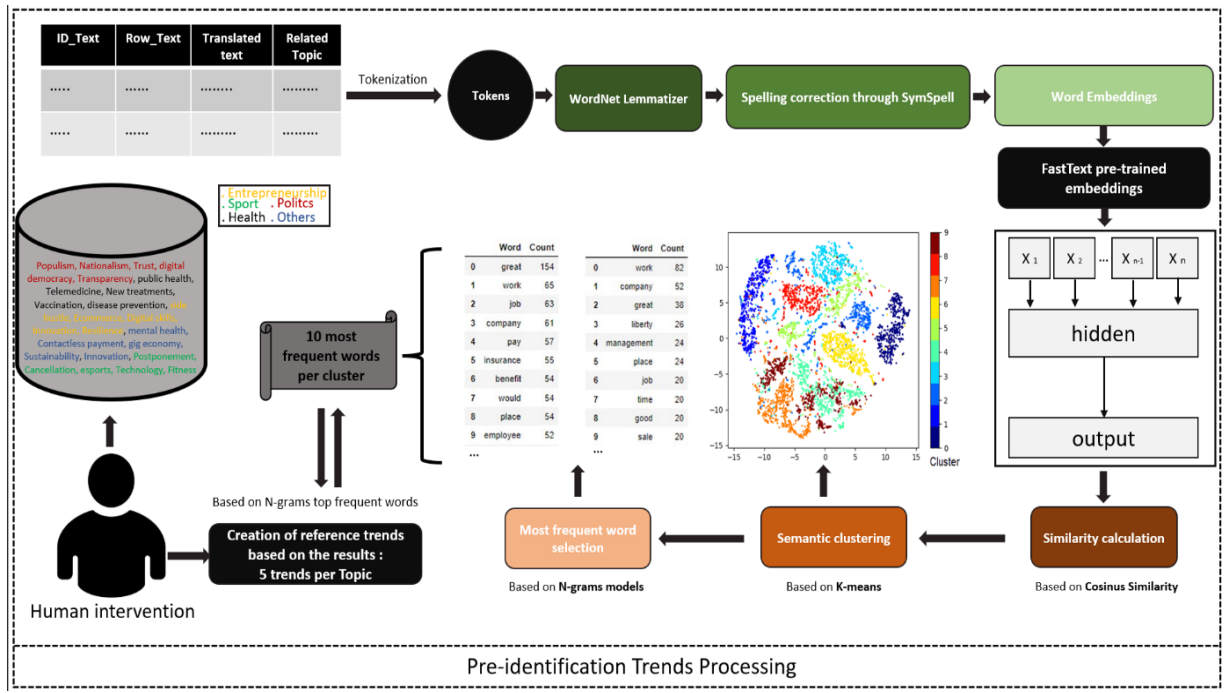

*Figure 4. The general architecture of the Pre-identification Trends Processing*

*Table 7. Word Embedding Statistics*

| Metric | Count/Percentage |
|---|---|
| Total Words | 4,000,831 |
| Unique Vectors Generated | 87,739 |
| Average Vector Dimension | 300 |
| Vocabulary Size | 89.234 |
| Out-of-Vocabulary Words | 1,495 |
| Sparsity of Vectors | 0.04 |
| Coverage of Vocabulary | 98.33% |

Following this stage, we begin the evaluation using the cosine similarity measure, a commonly used metric in natural language processing and information retrieval. It measures the similarity between two non-zero vectors by calculating the cosine of the angle between them, making it particularly beneficial for words categorization as it is not influenced by the magnitude of the vectors. This allows for the detection of high similarity between words with different term frequencies. Next, we employ K-means clustering for categorizing words based on their cosine similarities. Given a collection of words to be categorized, we compute the cosine similarity between the embedding vector of an uncategorized word and the embedding vectors of words already categorized. This is a pivotal step as it enables the vector quantization process, partitioning 4,000,831 observations into 10 clusters. During this process, each observation is allocated to the cluster with the nearest centroid, which acts as the cluster's prototype. This approach ultimately results in the formation of clusters consisting of words with the highest cosine similarities. Subsequently, within each of these word clusters, a 2-gram statistical model is employed to extract the 30 most frequent words. The 2-gram model, a specific type of probabilistic language model, predicts the next word in a sequence based on the previous word. This model computes the conditional probability of each word given its immediate predecessor in the sequence. The utility of using a 2-gram model in this context is its ability to capture not just the frequency of individual words, but also the frequency of word pairs, enhancing the understanding of contextual and semantic relationships between the words in the dataset.

Despite its robustness in capturing these relationships and trends, it is recognized that the 2-gram model may have limitations, particularly in capturing more nuanced meanings and trends that may not be fully represented by the frequency of word pairs alone. For this reason, the initial selection of words identified by the 2-gram model is further refined through manual intervention, where the selection is fine-tuned, ultimately reformulating five key trends for each topic based on the general context. Consequently, a total of 25 pre-defined trends, five from each of the five topics, are identified and serve as the reference trends for the entire dataset, as illustrated in Table 8.

*Table 8. Finalized Reference Trends Across Topics*

| Topic | Extracted Trends |
|---|---|
| Health | Public health, Telemedicine, new treatments, vaccination, Disease prevention |
| Sports | Postponement, Cancellation, E-sports, Technology, Fitness |
| Politics | Populism, Nationalism, Trust, digital democracy, Transparency |
| Entrepreneurship | Side Hustle, Ecommerce, Digital skills, Innovation, Resilience |
| Others | Mental Health, Contactless payment, gig economy, Sustainability, Education |

Lastly, using Temporal Component Analysis framework, the trend identification algorithm performs the essential task of correspondence and matching between the text and its final associated trend. This is accomplished by applying the AutoRegressive Integrated Moving Average (ARIMA) algorithm.

## 3.6. The ARIMA-Driven Final Trends Identification

Transitioning to the final methodology, as illustrated in Figure 5, we were initially faced with a significant obstacle since our dataset lacked a time dimension. To overcome this limitation, a step was undertaken by constructing an artificial time series. A time series comprises points indexed chronologically, typically observed at successive periods, making its analysis relevant to our study. This strategic move infused our dataset with a semblance of temporal continuity. For our purpose, texts were assigned sequential numerical identifiers, providing a foundational structure for this temporal analysis task. This method paved the way for the application of the ARIMA algorithm to unveil hidden temporal patterns within the textual content. Building upon this temporal foundation, the approach to identifying trends is firmly rooted in the ARIMA algorithm, represented as ARIMA$(p,d,q)$. In this context, "$p$" denotes the number of lag observations (autoregression) in the model, "$d$" represents the degree of differencing needed to make the time series stationary, Finally, "$q$" signifies the size of the moving average window, indicating the order of the moving average. The mathematical representation of the ARIMA model is:

$$X_t = \phi_1 X_{t-1} + \phi_2 X_{t-2} + \cdots \phi_p X_{t-p} + \theta_1 e_{t-1} + \theta_2 e_{t-2} + \cdots + \theta_q e_{t-q} + e_t \qquad (13)$$

where $X_t$ is the value of the series at time $t$, $\{\phi_1, \phi_2, \ldots, \phi_p\}$ are the parameters for the autoregressive terms, $\{\theta_1, \theta_2, \ldots, \theta_q\}$ are the parameters for the moving average terms, and $e_t$ is the white noise error term. The algorithm selection was grounded in an exhaustive model selection procedure that encompassed a spectrum of values for the parameters p, d, and q, and leveraged the AIC (Akaike Information Criterion) and BIC (Bayesian Information Criterion) values of various ARIMA models assessed during the hyper-parameter optimization phase. The AIC is a yardstick for assessing the quality of a statistical model. It is pivotal in model selection, wherein a model with a diminished AIC is generally favored over a model with a heightened AIC. The AIC appraises the model's quality by weighing both the model's goodness of fit and the quantity of parameters employed. Specifically, the AIC is computed as:

$$AIC = 2k - 2ln(L) \qquad (14)$$

where, $k$ represents the count of parameters in the model and $L$ symbolizes the maximized likelihood of the model. Analogous to the AIC, the BIC serves as an alternative criterion for model selection. It evaluates both the model's goodness of fit and the quantity of parameters utilized. However, it has a unique feature that imposes a stricter penalty on models with too many parameters compared to the AIC. Specifically, the BIC is computed as:

$$BIC = ln(n)k - 2ln(L) \qquad (15)$$

where, $(n)$ denotes the count of observations, $(k)$ represents the count of parameters in the model, and $(L)$ symbolizes the maximized likelihood of the model. A model with a diminished BIC is generally favored over a model with a heightened BIC. The BIC is especially valuable for comparing models with varying quantities of parameters, as it aids in averting overfitting by penalizing models with an excess of parameters. This process was an integral component during the grid search of hyper-parameters and was evaluated by the out-of-sample Root Mean Squared Error (RMSE) performance, highlighting its effectiveness in deciphering the complex temporal interdependencies inherent in time series data. The Figure 6 summarizes the optimal values of $p, d$ and $q$ for the ARIMA algorithm and their associated AIC and BIC values computed for each distinct topic.

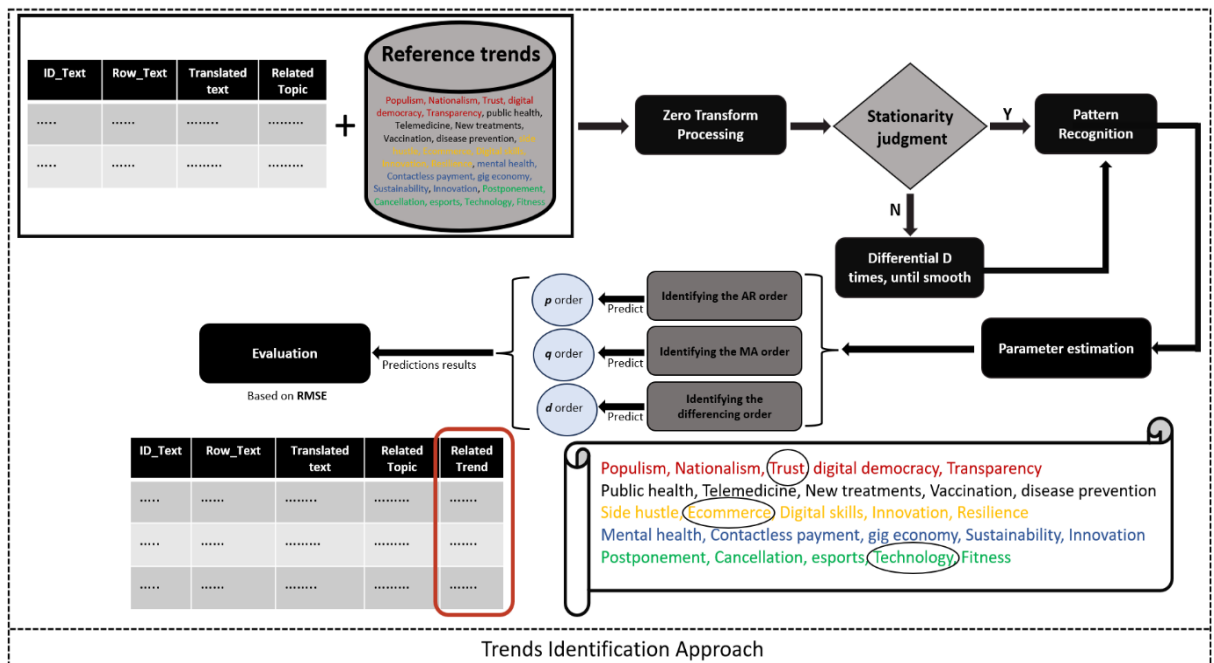

*Figure 5. The general architecture of the ARIMA-Driven Trends Identification*

As mentioned earlier, the ARIMA model is underpinned by three fundamental components: autoregression $(p)$, which captures the influence of past observations; differencing $(d)$, strategically applied to achieve stable stationarity and normalize variance; and moving average

($q$), adeptly leveraged to model historical forecast errors. The amalgamation of these components endows ARIMA with the capacity to extract valuable insights from time-dependent data. Furthermore, embarking on the ARIMA process entails a series of intricate steps that necessitate strict data preparation. Data values are transformed through differencing (d), a technique that bestows the desired level of stationarity onto the data. Before setting the (p, d, q) parameters, the process is diligently conducted based on the autocorrelation and partial autocorrelation functions with the primary objective of identifying the Autoregressive (AR) and Moving Average (MA) orders. This step, often termed as Pattern Recognition, ensures the model is finely tuned to the intrinsic temporal patterns within the data. In the parameter estimation step, the ARIMA model requires the identification of the AR, MA, and differentiation (d) orders, which are pivotal for producing accurate forecasts in time series data. Following this, each text line is precisely mapped to a specific trend, as predefined by the reference trends set established in the preceding phase. This process highlights the model's capability to decipher intricate temporal dynamics. Within this methodology, the intricate interplay between textual data and temporal trends is systematically elucidated, providing more accurate predictions and insights into topics and particularly addresses potential weaknesses when topics encompass overlapping subtopics such as the "Health" topic in our given study.

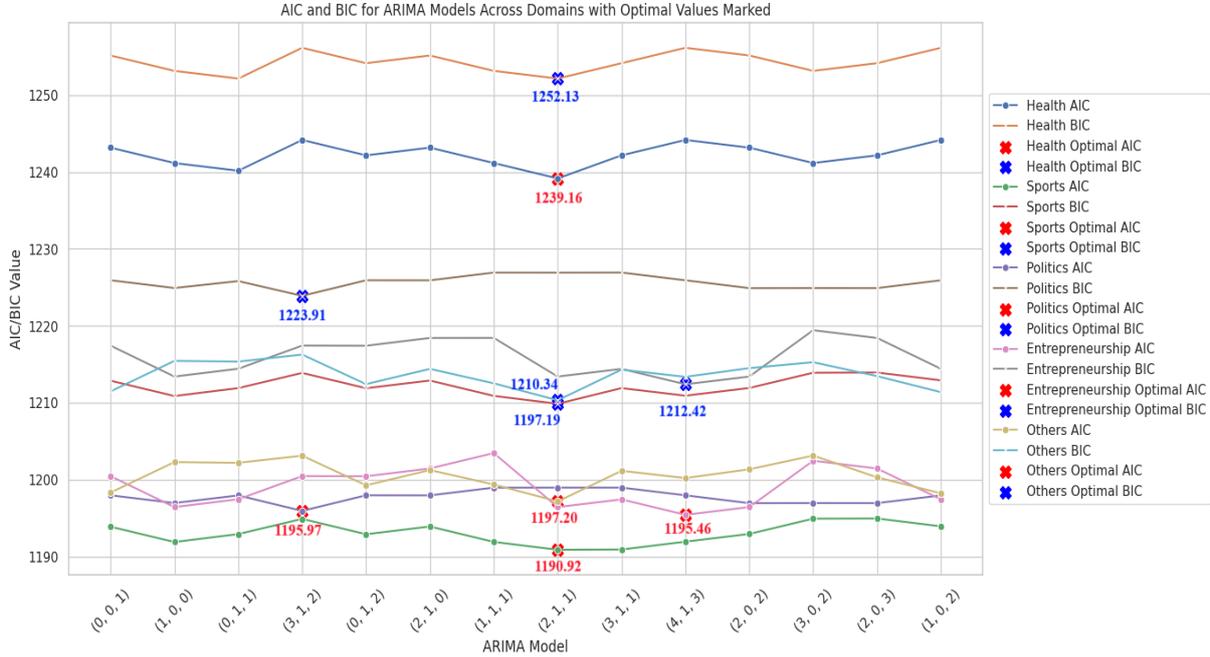

*Figure 6. Comparison of AIC and BIC Across Different ARIMA Models and Topics*

## 4. Experiments and Results

### 4.1. Baseline

In the experimental setup, it's essential to detail the methods and evaluation metrics used for assessing the different strategies applied in the study. This specific investigation focused on the evaluation of the three core approaches: translation, topic modelling, and trends identification. For the translation strategy, two pivotal metrics were employed: accuracy and F1-score. Accuracy quantifies the count of correct translations out of the aggregate translations executed by the model. Conversely, the F1-score is the harmonic mean of precision and recall, striking a balance between both. These metrics are instrumental in deciphering the effectiveness and efficiency of the developed translation model.

Regarding the topic modeling strategy, the C-score and U-mass scores were employed. The C-score quantifies the semantic coherence of the topics generated by the model. A superior C-score denotes that the words within a topic are more semantically interrelated. The U-mass score quantifies the similarity between topics, and a diminished U-mass score denotes that the topics are more divergent from one another. These metrics assist in evaluating the caliber and distinctiveness of the topics generated by the model.

Lastly, for the trend identification strategy, the RMSE metric was employed. RMSE quantifies the disparities between the values predicted by the model and the actual values. It offers an insight into the magnitude of the errors committed by the model in predicting trends. A diminished RMSE denotes a superior alignment of the model with the actual data. Table 9 provides the detailed formulas for all the above metrics, aiding in a comprehensive understanding of our evaluation setup.

*Table 9. Formulas for evaluation metrics used.*

| Metric | Formula |
|---|---|
| $Accuracy$ | $\dfrac{TP + TN}{TP + TN + FP + FN}$ |
| $F1 - Score$ | $\dfrac{2 \times TP}{2 \times TP + FP + FN}$ |
| $Score(w_i, w_j)$ | $\log\left(\dfrac{D(w_i, w_j) + 1}{D(w_i)}\right)$ |
| $C - score$ | $\sum_{i<j} score(w_i, w_j)$ |
| $U - mass$ | $\dfrac{1}{n(n-1)/2} \sum_{i<j} score(w_i, w_j)$ |
| $RMSE$ | $\sqrt{\dfrac{1}{n} \sum_{i=1}^{n} (y_i - \hat{y}_i)^2}$ |

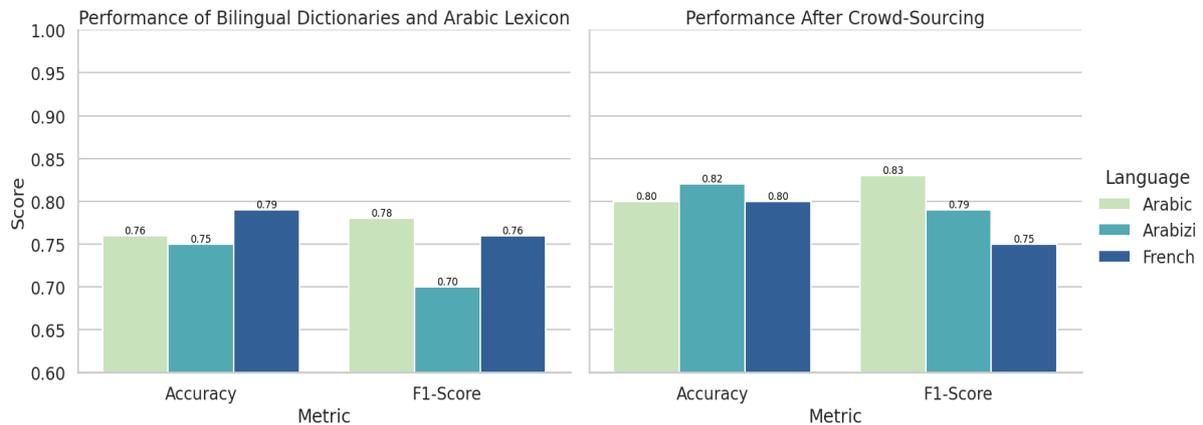

*Figure 7. Comparison of Translation Performance Metrics Before and After Crowd-Sourcing*

## 4.2. Results of First Experimental Approach

We begin our evaluation by inspecting the results of our translation approach, as visualized in Figure 7. The left bar chart offers a synopsis of machine-generated translations across three distinct languages: Arabic, Arabizi, and French. Collectively, the translations in Arabic and French showcased superior results in both accuracy and F1- score. On the contrary, Arabizi translations rendered intermediate results, while the right bar chart in the same Figure places emphasis on the transformative role of crowd-sourced translation in enhancing the efficiency of machine-generated translations. A marked increase in both accuracy and F1-score are observable post-human intervention. In this paradigm, Arabic translations reaped the most advantages, closely followed by those in Arabizi. This observation buttresses the notion that integrating a crowdsourced layer can substantially elevate the standard of automated translations. Our evaluation advances with Figure 8, a heatmap that shines light on the disparities in accuracy observed in machine-generated translations based on diverse data sources, each assessed individually before the collective merging of the three datasets. A discernible pattern emerges translation performance is not constant but fluctuates with variations in linguistic profiles and the intricacies of the data. CTSA corpus-sourced translations stood out, registering the highest accuracy across all datasets. Translations drawn from the CSTA and TUNIZI corpora, in contrast, delivered marginally subdued results. This reinforces the idea that the inherent character of data wields influence over the caliber of translations generated. In Graph 9, the data elucidates the correlation between the size of the Bilingual Dictionary and Translation Quality. An observable trend indicates that translation outcomes for languages, notably Arabic and French, are directly correlated with the depth and breadth of bilingual dictionaries employed. A further analysis suggests that this relationship is not incidental. As the scope of the dictionary expands, integrating an increased number of terms, idiomatic expressions, and linguistic intricacies, it provides a more extensive knowledge base for the translation algorithm. Consequently, this enrichment of the linguistic database facilitates the handling of an extensive spectrum of sentences, including those of higher complexity or idiosyncratic nature. The data thereby emphasizes the pivotal role that expansive linguistic repositories play in bolstering the precision and nuance of translation results.

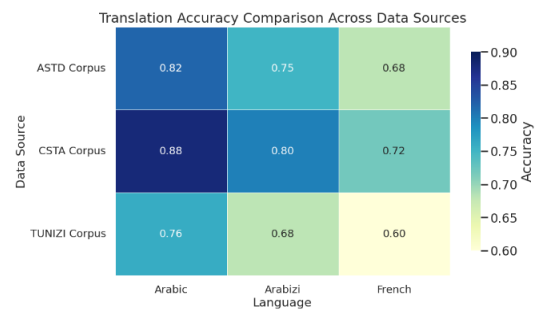

*Figure 8. Translation Accuracy Comparison Across Data Sources*

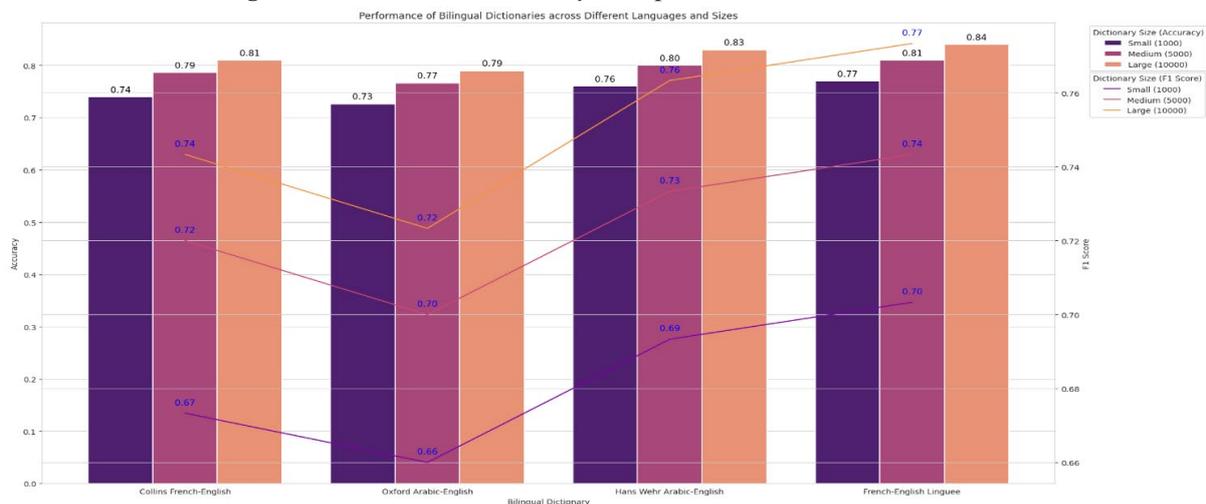

*Figure 9. Impact of Bilingual Dictionary Size on Translation Quality*

In turn, table 10 delineates the results derived from multiple iterations in which the integration of the OpenAI API was systematically employed to augment the precision of translations. Moreover, table 11 from our initial experimental results provides a qualitative analysis of some translations spanning Arabic, Arabizi, and French languages. It aims to contrast translations from our automated system with those from human translators.

*Table 10. Impact of OpenAI Api integration on Translation Accuracy*

| Iteration | Arabic | Arabizi | French |
|---|---|---|---|
| Initial Model | 0.76 | 0.75 | 0.79 |
| 1st Iteration | 0.78 | 0.78 | 0.81 |
| 2nd Iteration | 0.79 | 0.80 | 0.82 |
| 3rd Iteration | 0.81 | 0.80 | 0.84 |

*Table 11. Qualitative Analysis of System-Generated Translations vs. Human Interpretations*

| Source Text | Machine Translation system | Human-Translation |
|---|---|---|
| Na3rafek chkoun? | I recognize you, who is it? | I know you, who is it? |
| متى سنراك؟ | When will we see you? | When can we meet you? |
| Comment ça va? | How's it going? | How are you? |
| ماذا تفعل؟ | What are you up to? | What are you doing? |

## 4.3. Results of Second Experimental Approach

In the subsequent section, we commence our assessment by examining the results derived from our topic modeling methodology. Table 12 systematically presents the outcomes of hyperparameter optimization for both LDA and HDP models. The analysis of these tuning outcomes provides insights into the maximal performance potential of the respective models. For this assessment, we utilize perplexity and coherence as performance metrics. Conventionally, models with reduced perplexity and enhanced coherence values are favored, as they often yield more coherent and interpretable topics. This table provides a thorough insight into the algorithms' performance across varying configurations and facilitates the identification of the optimal hyperparameter settings. Within the comprehensive investigation of topic modeling methodologies, we have directed our emphasis specifically towards the HDP as a prominent technique under consideration. Notably, the metrics in Table 13 are categorized into two configurations: "HDP Set 1" and "HDP Set 2." While "HDP Set 1" is characterized by parameters (max_iter =100, alpha=0.1, beta=0.01, and gamma=5). Conversely, "HDP Set 2" operates under the settings (max_iter=150, alpha=0.05, beta=0.01, and gamma=10). Nevertheless, Table 14 elucidates the Comparison of Average Topic Modeling Metrics Across Distinct Models. This in-depth analysis underscores that, even with the commendable performance exhibited by the HDP, certain vulnerabilities persist, particularly stemming from the disparity of the five topics across the dataset.

*Table 12. Hyperparameter Tuning Results for LDA And HDP Models*

| Model | Alpha | Beta | Gamma | Perplexity | Coherence |
|---|---|---|---|---|---|
| LDA | 0.01 | 0.01 | -- -- | 175.2 | 0.55 |
| LDA | 0.05 | 0.05 | -- -- | 174.5 | 0.56 |
| LDA | 0.01 | 0.1 | -- -- | 177.1 | 0.54 |
| LDA | 0.1 | 0.01 | -- -- | 178.0 | 0.53 |
| LDA | 0.025 | 0.025 | -- -- | 176.2 | 0.55 |
| LDA | 0.075 | 0.075 | -- -- | 175.4 | 0.56 |
| LDA | 0.03 | 0.03 | -- -- | 176.5 | 0.55 |
| LDA | 0.06 | 0.06 | -- -- | 175.8 | 0.56 |
| LDA | 0.04 | 0.04 | -- -- | 176.0 | 0.55 |
| HDP | 0.1 | 0.01 | 5 | 95 | 0.57 |
| HDP | 0.2 | 0.01 | 5 | 100 | 0.52 |
| HDP | 0.3 | 0.01 | 5 | 101 | 0.50 |
| HDP | 0.1 | 0.001 | 5 | 98 | 0.56 |
| HDP | 0.2 | 0.001 | 5 | 99 | 0.54 |
| HDP | 0.3 | 0.001 | 5 | 100 | 0.52 |
| HDP | 0.1 | 0.01 | 10 | 98 | 0.57 |
| HDP | 0.2 | 0.01 | 10 | 99 | 0.55 |
| HDP | 0.5 | 0.01 | 10 | 98 | 0.56 |
| HDP | 0.1 | 0.005 | 5 | 99 | 0.55 |
| HDP | 0.2 | 0.005 | 5 | 100 | 0.53 |
| HDP | 0.5 | 0.005 | 5 | 99 | 0.54 |
| HDP | 0.1 | 0.0005 | 5 | 98 | 0.57 |
| HDP | 0.2 | 0.0005 | 5 | 99 | 0.55 |
| HDP | 0.5 | 0.0005 | 5 | 98 | 0.56 |

*Table 13. C-score and U-mass Scores for Topics Using HDP Topic Modeling*

| Topic | Model | C-score | U-mass |
|---|---|---|---|
| Entrepreneurship | HDP Set 1 | 0.74 | 0.85 |
| Sport | HDP Set 1 | 0.65 | 0.74 |
| Politics | HDP Set 1 | 079 | 0.80 |
| Health | HDP Set 1 | 0.70 | 0.75 |
| Others | HDP Set 1 | 0.60 | 0.65 |
| Entrepreneurship | HDP Set 2 | 0.73 | 0.75 |
| Sport | HDP Set 2 | 0.62 | 0.65 |
| Politics | HDP Set 2 | 0.71 | 0.75 |
| Health | HDP Set 2 | 0.65 | 0.64 |
| Others | HDP Set 2 | 0.52 | 0.55 |

*Table 14. Comparison Analysis of Average Topic Modeling Metrics Across LDA And HDP*

| Model | Avg C-score | Avg U-mass | Avg Topic Coherence |
|---|---|---|---|
| HDP Set 1 | 0.73 | 0.82 | 0.83 |
| LDA | 0.68 | 0.77 | 0.79 |
| HDP Set 2 | 0.75 | 0.85 | 0.86 |
| LDA | 0.71 | 0.76 | 0.80 |

*Note: The Gamma parameter is specific to HDP and does not apply to LDA.

Given that the primary objective of this study is to validate the concept of predicting topics, figure 10 illustrates the proportions of the five topics over the initial 500 timestamps, accompanied by their moving averages. Notably, certain topics are trending upwards (e.g., Health and Sport), some downwards, while others demonstrate random fluctuations across the analyzed timestamps (e.g., Entrepreneurship)

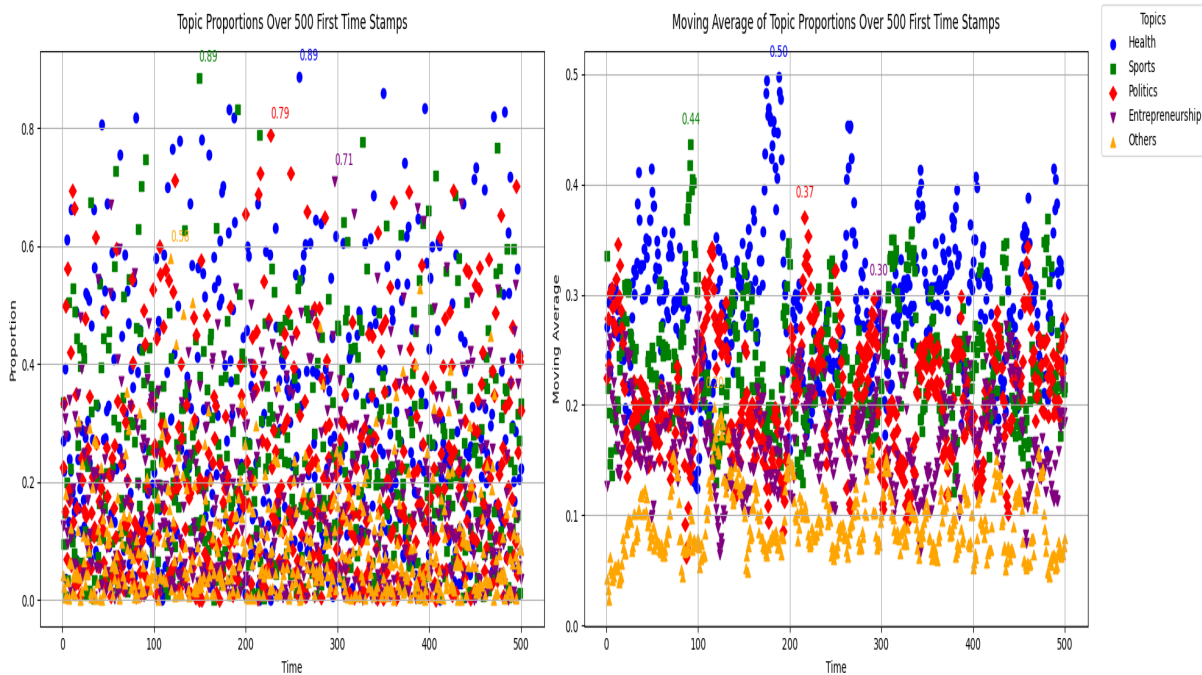

*Figure 10. Temporal Evolution of Topic Proportions Over the First 500 Timestamps with Moving Averages*

## 4.4. Results of Third Experimental Approach

In our third experimental approach, we methodically evaluated the performance of the ARIMA-Driven Final Trends Identification, focusing primarily on two aspects: its capacity for out-of-sample forecasting and the validation of hyperparameter optimization. The primary objective concerning out-of-sample forecasting was to rigorously assess the model's predictive accuracy on data subsets not encountered during its training. For this purpose, 70% of the available data was used to train the model, leaving the remainder as an untouched test set or out-of-sample data. Post-training, this model was used to generate forecasts on this untouched data. Subsequent analysis involved the calculation of the Root Mean Squared Error (RMSE), a metric gauging the discrepancies between the predicted and actual values from the test dataset, given that a lower RMSE typically suggests enhanced forecasting precision. Turning to hyperparameter optimization validation, the primary objective was to identify the ideal parameter constellation for the ARIMA model, ensuring a balance between statistical precision and ease of interpretation. This involved an extensive grid search across varied values for the parameters p, d, and q. Simultaneously, the model's AIC and BIC scores were monitored for each parameter combination, as illustrated in Figure 6 above. These scores act as pivotal indicators, gauging the model's alignment with the data and imposing penalties for undue complexity. To strengthen the validation of our hyperparameter selections, we employed correlation methods. As depicted in Figure 11, both AIC and BIC scores for various ARIMA models across different topics are plotted. The optimal configurations are distinctly highlighted using the 'X' symbol in the graph, which signifies the best model configuration for each topic based on both AIC and BIC values. Through our detailed analysis, it became clear that the ARIMA setups of (2,1,1), (3,1,2), and (4,1,3) were the standout choices.

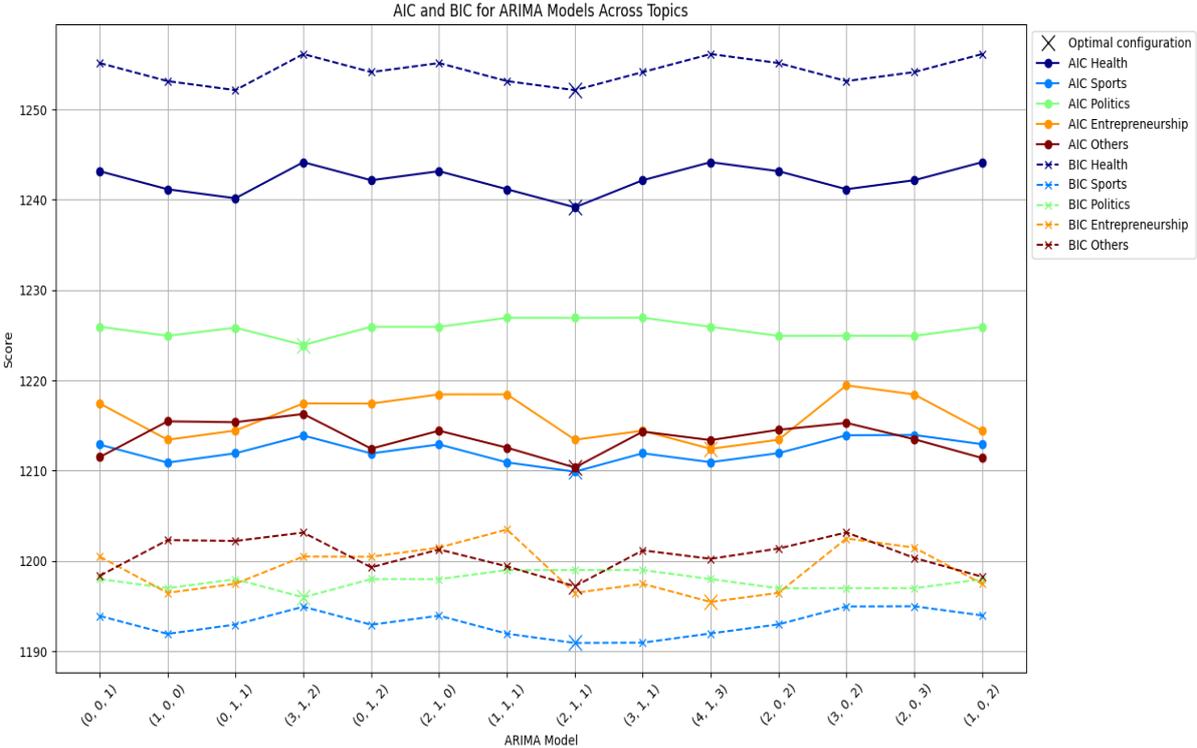

*Figure 11. Optimal Configurations of ARIMA Models via AIC & BIC Scores Across Topics*

Furthermore, Figure 12 provides a 3D diagram illustrating the interplay between trends, ARIMA configurations, and RMSE values, offering a lucid representation of performance metrics across the 25 trends. The X-axis distinctly portrays trends, showcasing categories such as "Populism", "Digital Democracy", "Public Health", and "Ecommerce". The Y-axis demystifies the optimal ARIMA configurations for each trend which are encapsulated in the triad (p, d, q). The Z-axis, represented by the varying heights of bars, reflects the RMSE values, shedding light on the alignment between the model's forecasts and the actual outcomes.

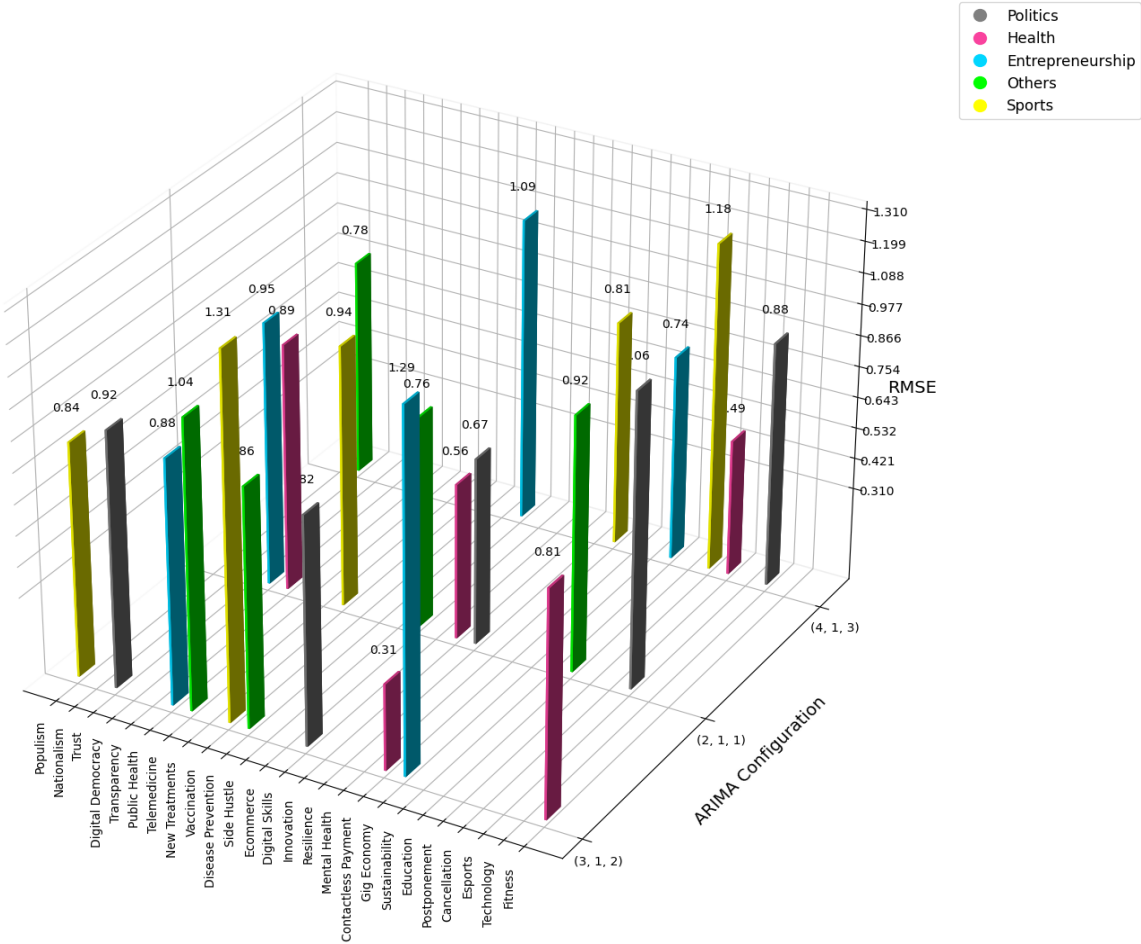

Figure 12. Trends Forecasting by Optimal ARIMA Configurations Across Topics

In table 15, we present the top five trends as determined by their respective RMSE scores, with each row representing a specific trend ascertained from an ARIMA model. The trends have been systematically organized in the table in ascending order of their RMSE values. It warrants emphasis that a lower RMSE score suggests enhanced accuracy and reliability of the corresponding model. As such, the trend situated at the foremost position in the table is deemed to have the most accurate forecast as per its ARIMA model configuration.

Table 15. Top five trends ranked by forecasting accuracy using ARIMA model RMSE score

| Trend | ARIMA (p,d,q) | RMSE | Related Topic |
|---|---|---|---|
| Public Health | (2, 1, 1) | 0.31 | Health |
| Telemedicine | (4, 1, 3) | 0.49 | Health |
| New Treatments | (3, 1, 2) | 0.56 | Health |
| Populism | (3, 1, 2) | 0.67 | Politics |
| Side Hustle | (4, 1, 3) | 0.74 | Entrepreneurship |

## 4.5. Discussion

In this section, we will distill the most significant conclusions and discuss the results of each approach, respectively: the translation strategy, the topic modeling approach, and the trends identification process. Additionally, we will critique the weaknesses inherent to each method.

Initially, our approach delved deep into machine translation capabilities, emphasizing three dominant languages: Arabic, Arabizi, and French. Through this lens, we engaged with a multitude of analytical perspectives, aiming for a comprehensive grasp of the subject. One standout observation from Figure 7 highlighted that Arabic and French (about 76% and 79% accuracy), given their standardized nature and extensive documentation, consistently delivered commendable translation accuracy. However, Arabizi, straddling a gray area linguistically, presented more subdued results, as further clarified by Figure 8, reaching around 68% in the worst cases. Moving on, when we integrated human expertise from the ProZ Platform, the translation process underwent a marked transformation, with Arabic translations particularly benefiting. This shift shed light on the invaluable role human insight plays in honing machine-generated outcomes. Subsequently, our exploration of Figure 8 unveiled discernible variations in machine translation results, dependent on the origins of each individual dataset. In particular, the CTSA database-backed translations surpassed their counterparts in accuracy (varies from 72% to 88%), reinforcing the profound impact of dataset integrity on translation quality. Transitioning to Figure 9, the essence of a meticulously compiled bilingual dictionary for prime translation was accentuated. The graph manifested a direct relationship between the scope of a dictionary and the ensuing translation quality. Further validated by the same figure, broader dictionaries empower translation models to discern and render delicate linguistic subtleties and nuances inherent to their respective target languages. Hence, we strongly recommend utilizing dictionaries spanning a medium to extensive range, ensuring they align with the distinct characteristics of every language duo (i.e., + 6% improvement with the Oxford Arabic-English Dictionary). Shifting our focus to Table 10, the benefits of subsequent iteration through the OpenAI API became evident. The initial model registered moderate accuracy, but with each iteration, there was a clear improvement in results. The initial model exhibited only modest accuracy levels, but with each subsequent iteration, there were noticeable improvements in the results. This progressive trend strongly demonstrates the effectiveness of the OpenAI API in improving machine translation quality. In conclusion, it is crucial to acknowledge the variations in the extent of translation enhancements among the languages studied. Notably, the integration of the API yielded more significant benefits for the accuracy of Arabic translations than for Arabizi or French. Such variances may be attributed to the inherent linguistic complexities associated with each language, and how the machine translation models are specifically designed to handle these nuances. However, these disparities underscore a potential limitation: the model's unequal adeptness across different languages, potentially stemming from insufficient training data or lack of representativeness for certain languages. Reflecting on our study, we have garnered insights into the multifaceted mechanics of machine translation. While elements such as human expertise, and careful dataset choices are pivotal in steering superior translation outcomes, it's crucial to recognize that their efficacy might vary based on the particularities of the language in question.

Embarking on our comprehensive examination of topic modeling methodologies, with particular focus on LDA and HDP, we observed nuanced responses to hyperparameter adjustments in both models as illustrated in Table 12. Optimal hyperparameter configurations were shown to enhance topic coherence and reduce perplexity, subsequently producing more interpretable topics. A deeper dive into the results highlights that certain hyperparameter combinations, especially around Alpha: 0.05, Beta: 0.05 and Alpha: 0.075, Beta: 0.075, seem favorable for the LDA model. These combinations appear to promote a balanced distribution of topics and words within documents, heightening coherence. Conversely, for the HDP model, while the gamma hyperparameter's impact on coherence remained marginal, notable improvement was evident with higher alpha and beta values, especially around {Alpha: 0.1, Beta: 0.01 and Alpha: 0.1, Beta: 0.005}. Transitioning to the metrics from Table 13, we identified significant variability in C-scores and U-mass scores across different topics. Topics such as Politics in "HDP Set 1" secured remarkably high C-scores, while others, like the "Others" category in "HDP Set 2", trailed. This variability hints that some topics might inherently lend themselves more fittingly to the HDP modeling approach. Moreover, variations in scores between the two HDP sets for identical topics indicate the model's susceptibility to changes in hyperparameter configurations or dataset divides. For instance, the topic of "Entrepreneurship" underwent a noticeable drop in both C-score and U-mass transitioning from Set 1 to Set 2. An intriguing observation, though, is the equivalence of U-mass scores to C-scores for the same topic, insinuating the model's prowess in distinguishing between different topics. Further evaluation of Table 14, which presents a side-by-side comparison of average topic modeling metrics between LDA and HDP, unraveled several discernible patterns. HDP, in both sets, consistently surpassed LDA in all metrics, suggesting its relative superiority in extracting coherent and distinct topics, at least within the confines of this study. An underlined trend across the board was the synchronous movement of metrics; when one metric rose, the others typically followed suit, suggesting a possible interrelation among coherence, distinctiveness, and overall topic quality. Notably, while "HDP Set 2" slightly edged out Set 1 in all metrics, the performance gap between LDA and HDP shrank from Set 1 to Set 2, emphasizing that while HDP may lead the charge, LDA remains a competent contender, with its effectiveness potentially hinging on specific dataset traits or configurations. Nonetheless, HDP's inconsistencies in certain configurations insinuate the need for model refinement or a revisit to its core architecture. Relying exclusively on quantitative metrics like perplexity and coherence might not yield a holistic understanding; thus, future endeavors could benefit from roping in human evaluators to gauge the topics' intuitive clarity and relevance. The conundrum remains: are topic fluctuations attributed to external factors or intrinsic model mechanics? This study brings forth two pivotal insights: the reaffirmation of hyperparameter fine-tuning's significance, and the innovative approach of analyzing topic proportions temporally, providing insights for entities keen on monitoring shifting public sentiments. Turning our attention to Figure 10, which delineates the proportions of our five topics over the first 500 timestamps, enriched with moving averages, we observe a kaleidoscope of trends. Specifically, the "health" and "sport" topics conspicuously dominate in proportions, underscoring their sustained relevance. Conversely, the "others" and "entrepreneurship" topics, both in direct readings and moving averages, reveal more subdued patterns, hinting at their periodic prominence or perhaps a broader encapsulation of varied content, especially within the "others" bracket. This disparity possibly mirrors a confined audience interest or the dataset's inherent biases.

Progressing with the analysis derived from our trend identification methodology, a detailed scrutiny of both Figure 6 and Figure 11 offers revealing insights. Specifically, the optimal ARIMA configurations for various topics are discernibly associated with the minimized combination of AIC and BIC values. Expanding on this observation: In the health domain, the ARIMA (2, 1, 1) model exhibits a superior AIC value of 1239.160, complemented by a competitive BIC. For sports-related data, both ARIMA (1, 0, 0) and ARIMA (2, 1, 1) configurations showcase low AIC metrics. However, a combined evaluation of AIC and BIC suggests that ARIMA (2, 1, 1) is more optimal. In the political arena, the ARIMA (3, 1, 2) model marginally outperforms its counterparts based on AIC. Within the entrepreneurship sector, the ARIMA (4, 1, 3) configuration demonstrates a noteworthy AIC of 1195.460 and a BIC of 1212.418. For other subjects, the ARIMA (2, 1, 1) configuration is preeminent with an AIC of 1210.345. Nevertheless, one must be cautious while interpreting these results. A foundational assumption of the ARIMA model is the linearity of the data. Given the multifaceted nature of the studied topics, potential nonlinearities may exist. Additionally, ARIMA models can be sensitive to abrupt shifts in data patterns. Consequently, major global events could significantly influence the forecasting accuracy of the model. This underscores the necessity of considering alternative modeling techniques or hybrid methodologies, particularly for volatile data series. The results provide compelling evidence of the ARIMA model's adaptability across diverse subjects. The model's capacity to accommodate varied configurations elucidates its robustness. Yet, this flexibility calls for meticulous model selection, especially in datasets characterized by instability. Utilizing both AIC and BIC metrics concurrently proves instrumental in this regard. Additionally, the 3D graph (figure 12) serves as a testament to the prowess of trend forecasting via the ARIMA model. Displayed trends encompass both socio-political vectors such as "Populism" and "Digital Democracy" and health-centric domains including "Public Health" and "Telemedicine". These diverse trends offer insights into evolving societal dynamics. Diving deeper into the ARIMA model's specifics, each trend's predictability is contingent upon its unique ARIMA configuration, characterized by the parameters (p, d, q). The heterogeneity of these configurations reiterates the data's intrinsic variability and the tailored approach requisite for optimal results. Furthermore, a notable observation is the "Public Health" trend, with an RMSE value of 0.31, suggesting a high degree of predictive accuracy. Conversely, elevated RMSE values in other trends (i.e., about 1.18 RMSE value with "Postponement" trend) indicate avenues for model refinement and the potential adoption of alternative analytical methods. On a broader scale, trends are systematically categorized into overarching domains. This stratification provides a structured perspective of prevailing societal themes, imparting insights to a diverse spectrum of stakeholders, ranging from policymakers, instrumental in guiding pandemic management strategies, to the academic community.

## 5. Conclusion

In this paper, we delved deeply into the multifaceted domain of multilingual topic dynamics, with a concentrated lens on the Tunisian textual production during the Coronavirus pandemic.

A multi-faceted approach was adopted, bridging machine translation, advanced topic modeling, and temporal trend identification. The study underscored the importance of dataset integrity, linguistic expertise, and hyperparameter optimization in yielding coherent and representative results.

The study's findings pointed to the commendable accuracy achieved in translating standardized languages such as Arabic and French. In contrast, the adaptive characteristics of Arabizi posed inherent challenges. The dynamics between HDP and LDA in topic modeling were stark, their performance being closely aligned with chosen hyperparameters and the specificities of the datasets. Additionally, our application of the ARIMA model reaffirmed its mettle in tracking temporal trends, yet its susceptibility to sudden data perturbations highlighted the imperative for meticulous utilization.

Our future works will focus on improving our understanding of topic modeling, given the noticeable inconsistencies in metrics. We're considering hybrid topic modeling and neural topic models for deeper insights. Additionally, we aim to enhance the temporal analysis methods, especially with the ARIMA model's limitations for volatile data, looking at tools like Prophet and Combined Bi-LSTM.

**Acknowledgment**. The authors would like to express their gratitude to the members of RLANTIS Lab (University of Monastir) and LaBRI Lab (University of Bordeaux) for their significant assistance and contributions in the preparation of this paper.

**Data Availability Statement.** The primary corpora and the associated translated dataset from this study are available upon request from the corresponding author.